\documentclass[sigconf]{acmart}

\AtBeginDocument{%
  }

\copyrightyear{2026}
\acmYear{2026}
\setcopyright{cc}
\setcctype{by}
\acmConference[WWW '26]{Proceedings of the ACM Web Conference 2026}{April
13--17, 2026}{Dubai, United Arab Emirates}
\acmBooktitle{Proceedings of the ACM Web Conference 2026 (WWW '26), April
13--17, 2026, Dubai, United Arab Emirates}
\acmPrice{}
\acmDOI{10.1145/3774904.3792991}
\acmISBN{979-8-4007-2307-0/2026/04}

\newcommand{\method}{\textsc{VLTCrisis}}
\newcommand{\dataset}{\textsc{CrisisMMD}}
\usepackage{multirow}
\usepackage{soul}
\usepackage{color}
\usepackage{tabularx}
\usepackage{amsfonts}
\usepackage{amsmath}
\newcolumntype{S}{>{\hsize=.45\hsize}X}
\newcolumntype{L}{>{\hsize=.55\hsize}X}
\newcolumntype{W}{>{\hsize=.70\hsize}X}
\newcolumntype{M}{>{\hsize=.15\hsize}X}
\newcolumntype{T}{>{\hsize=.12\hsize}X}

\begin{document}

\title{Cross-Modal Rationale Transfer for Explainable Humanitarian Classification on Social Media}

\author{Thi Huyen Nguyen}
\affiliation{%
  \institution{L3S Research Center \\Leibniz University Hannover}
  \city{Hannover}
  \country{Germany}}
\email{nguyen@l3s.de}

\author{Koustav Rudra}
\affiliation{%
  \institution{Indian Institute of Technology Kharagpur}
  \city{Kharagpur}
  \country{India}}
  \email{krudra@ai.iitkgp.ac.in}

\author{Wolfgang Nejdl}
\affiliation{%
 \institution{L3S Research Center\\Leibniz University Hannover}
 \city{Hannover}
 \country{Germany}}
 \email{nejdl@l3s.de}






\begin{abstract}
 Advances in social media data dissemination enable the provision of real-time information during a crisis. The information comes from different classes, such as infrastructure damages, persons missing or stranded in the affected zone, etc.
Existing methods attempted to classify text and images into various humanitarian categories, but their decision-making process remains largely opaque, which affects their deployment in real-life applications. Recent work has sought to improve transparency by extracting textual rationales from tweets to explain predicted classes. However, such explainable classification methods have mostly focused on text, rather than crisis-related images. In this paper, we propose an interpretable-by-design multimodal classification framework. Our method first learns the joint representation of text and image using a visual language transformer model and extracts text rationales. Next, it extracts the image rationales via the mapping with text rationales. Our approach demonstrates how to learn rationales in one modality from another through cross-modal rationale transfer, which saves annotation effort. Finally, tweets are classified based on extracted rationales. Experiments are conducted over \dataset{} benchmark dataset, and results show that our proposed method boosts the classification Macro-F1 by 2-35\% while extracting accurate text tokens and image patches as rationales. Human evaluation also supports the claim that our proposed method is able to retrieve better image rationale patches ($12\%$) that help to identify humanitarian classes. Our method adapts well to new, unseen datasets in zero-shot mode, achieving an accuracy of 80\%. 
\end{abstract}






\maketitle

\section{Introduction}
\label{sec:intro}


Users post lots of multimodal information (e.g., text, image) during crisis situations.  
Efficient classification tools have been investigated to identify useful content for situation awareness and relief efforts.
Recent studies proposed many text- and image-based methods~\cite{liu2021crisisbert,alam2022robust,shetty2024disaster}  that tried to identify humanitarian classes (e.g., infrastructure damage, affected persons, volunteer operations, etc.) or assess the needs of victims, and damage quality (e.g., severe/mild/low), etc. In multimodal analysis, some contents are complementary in nature, i.e., contents of one modality help cover the missing information in another modality. 
Most of the existing methods focused on optimizing model performance. Recent works focused on interpretable crisis-related microblog classification and summarization~\cite{nguyen2022rationale,nguyen2022crisicsum}. However, the authors only explored textual content. 
There were also works on image interpretability, but most of them relied on posthoc strategies. 
In recent times, multimodal interpretability has gained attention because multimodal approaches are applied in different sensitive domains such as autonomous driving, healthcare, etc. The United Nations Development Programme also states the requirement of transparency in the design and use of machine learning models for humanity~\cite{UNDP}. Researchers tried to address the interpretability problem in various ways~\cite{joshi2021review,liang2022multiviz} based on attention, counterfactual, etc.
However, the problems are --- (i).~ Such methods are primarily posthoc in nature rather than interpretable by design, (ii).~Though some interpretable by design approaches exist for text~\cite{zhang2021explain,nguyen2022rationale}; they are difficult to adapt for images, (iii).~Existing works did not try zero-shot rationale transfer across different modalities.


\begin{figure}[tb]
    \centering
    \includegraphics[scale=0.42]{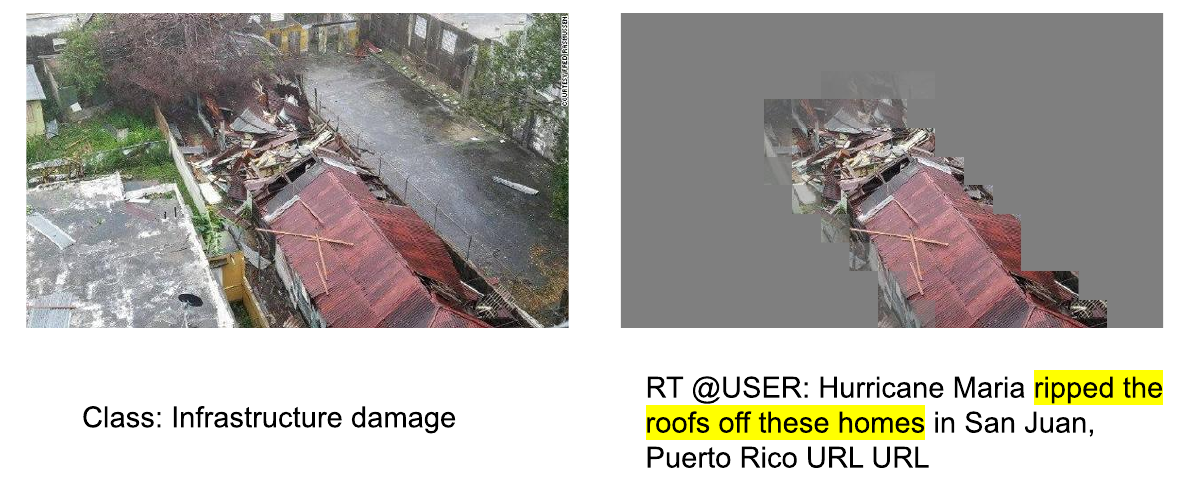}
    \vspace{-4mm}
    \caption{An example of a tweet about `infrastructure damage'. Highlighted words and unmasked image patches are rationales.}
    \label{fig:example}
    \vspace{-4mm}
\end{figure}

In this paper, our focus is on multimodal content (text and images) and developing an \textit{interpretable by design} multimodal classification framework for crisis-related tweets.
Specifically, we consider the humanitarian class detection task of tweets containing both text and image. Along with identifying the class, the objective is to find explanations from both text and image supporting the model's decision. 
Figure~\ref{fig:example} shows an example of such a scenario. Our proposed \textbf{V}ision and \textbf{L}anguage \textbf{T}ransformer-based approach for \textbf{Crisis}-related tweet classification method (\method{}) works in two stages. First, tweets (texts and images) are passed through a visual-language transformer model (ViLT)~\cite{kim2021vilt} to learn their representations. The encoder also considers the interaction between text and image components. Next, we apply a decoder layer that solves three tasks based on the shared representation: (i). identifying tweet class, (ii). predicting text rationales, (iii). learning image rationale heatmap by aligning text rationales to image patches. This transfer learning helps to obtain image rationales directly from text counterparts without any human annotation of image patches. After extracting text and image rationales, we pass them through ViLT and use the learned representations to predict class labels of tweets in the second stage. This step ensures interpretability by design, as the prediction of tweet labels is solely based on extracted rationales. Our \method{} works on the assumption that the text is aligned/correlated with the image, i.e., whatever is mentioned in the text is supported by the image and vice versa. 
While this assumption may not always hold in general social media, our empirical validation (Section \ref{subsec:alignment}) shows that crisis-related posts exhibit strong alignment on average. 
Extensive experiments conducted over a multimodal dataset -- \dataset{}~\cite{alam2018crisismmd} show the efficacy of our method. The contributions of our paper are as follows: 

\begin{itemize}
    \item We propose an interpretable approach for the multimodal classification of tweets during crisis events\footnote{The code will be made publicly available at \url{https://github.com/thi-huyennguyen/VTLCrisis}}. 
    \item We annotate the text rationales and generate image rationales through transfer learning. We show that it is possible to learn rationales of one modality from another. This significantly reduces the annotation effort.
    \item We illustrate the role of texts and images for classification in our learning approach. The joint learning strategy helps align the concepts across modalities. Text and image complement each other, yielding better ($5-6\%$) representations than individual modalities.
    \item Experiments show that our interpretable multimodal classification approach achieves competitive or better (2-35\% Macro-F1) performance than existing methods.
    \item Our proposed model is highly faithful and plausible. The model performance drops by $51\%$ in the absence of identified rationales.
    \item Our proposed model adapts well to new, unseen multimodal crisis datasets in zero-shot mode, yielding approximately $ 80\%$ accuracy. 
\end{itemize}


\section{Related work}
\label{sec:related}
\subsection{Tweet Classification}
\noindent\textbf{Unimodal Classification:} Crisis events create lots of disruption in affected zones. Existing methods tried to use different modalities such as text and image~\cite{rudra2018identifying,liu2021crisisbert,nguyen2023learning,sanchez2023cross}, etc. Research studies relied upon various approaches such as bag-of-words~\cite{aidr2014}, matrix factorization~\cite{li2018disaster,mazloom2019hybrid}, neural models~\cite{DatNguyen2017,neppalli2018deep}, etc. Recently, some studies evaluate the capacities of instruction-finetuned Large Language Models (LLMs) for the classification of crisis-related tweets~\cite{mcdaniel2024zero,yin2024crisissense}. 

Some works~\cite{alam2018processing} proposed an image processing pipeline to detect damage severity. These take care of noise cleaning, relevant real-time image streaming, and damage assessment. Further, Li~et~al~\cite{li2019identifying} proposed a domain adversarial neural network (DANN) to perform domain adaptation between source and target crises. Li~et~al~\cite{li2019localizing} used class activation mapping to identify damaged regions and the severity of damage. Recent works~\cite{long2023crisisvit} relied on transformer-based algorithms for crisis image classification.\\

\noindent\textbf{Multimodal Classification:} Some recent studies~\cite{alam2018crisismmd,hao2020leveraging} pointed out the limitations of individual modalities, i.e., text or images, and the benefits of multimodal data for crisis analysis. Alam~et~al.~\cite{alam2018crisismmd} introduced a benchmark multimodal dataset for the crisis domain, which enabled a range of methods to address humanitarian classification more effectively~\cite{mouzannar2018damage,said2019natural,rizk2019computationally}. Abavisani~et~al.~\cite{abavisani2020multimodal} proposed a cross-modal architecture to mitigate misleading signals from weak modalities. Subsequent work focused on further improving classification performance through enhanced fusion and representation learning~\cite{gautam2019multimodal,ofli2020analysis,basit2023natural,shetty2024disaster,ma2025camo}. More recently, large language models have been explored for crisis tweet classification~\cite{yin2024crisissense}. However, fine-tuning LLMs is computationally expensive and data-hungry, and the relatively small size of crisis-related multimodal datasets increases the risk of overfitting compared to lightweight, task-specific models. Moreover, these approaches treat multimodal models as black boxes with limited insight into decision evidence.

\subsection{Model Interpretability}
Interpretability is incorporated in three different phases of modeling --- (i). Pre-modeling (ii). In-modeling, and (iii). Post-modeling. Pre-modeling approaches deal with data cleaning, noise removal, attribute selection, etc. Posthoc explanations could be grouped into various categories based on explanation scope and model accessibility~\cite{liu2022trustworthy}. As per model accessibility, the methods could be {\it model agnostic}~\cite{lakkaraju2019faithful} or {\it model intrinsic}~\cite{qi2019visualizing}. As per the scope of explanation, we have local and global explanation methods. Local explanation methods (e.g., LIME~\cite{ribeiro2016should}, SHAP~\cite{lundberg2017unified}) tried to explain the prediction of a single instance. Global explanation methods~\cite{gacto2011interpretability} give an overall explanation of the model. However, such approaches could be easily fooled~\cite{slack2020fooling}. In-modelling approaches have gained a lot of attention in recent times. Decision trees~\cite{optimal2023shati}, rule-based systems~\cite{lakkaraju2016interpretable,ghosh2023interpretability} are inherently interpretable. Recent studies proposed {\it interpretable by design} models that tried to learn both tasks and explanations simultaneously~\cite{DeYoung2020,zhang2021explain}.

Some studies provide a detailed survey on black-box interpretable models for text or image data~\cite{choudhary2022interpretation,chan2022unirex,kamakshi2023explainable}. 
The majority of previous works rely on posthoc explanation approach. Despite various studies in this direction, there is still no consensus on how to evaluate model interpretability due to the lack of groundtruth for posthoc evaluation.  
Recently proposed visual transformers~\cite{kim2022vit,liu2023survey} bring significant changes in image-related tasks. A few studies~\cite{al2023interpretable,wang2023visual,parekh2024concept} propose methods on multi-modal model interpretability. However, these methods face challenges when applying to humanitarian classification problems due to the unstructured and complex association of image-text pairs from microblogs.

\subsection{Interpretable methods in crisis domain}
Most of the existing crisis-related classification methods focus on predictive performance~\cite{aidr2014,rudra2018identifying}. CrisisKAN~\cite{gupta2024crisiskan} introduces a knowledge-infused attention network, where external crisis knowledge guides multimodal fusion and improves interpretability through attention weights. Recent methods focused on rationale-based interpretability of the models along with the performance~\cite{Nguyen2022towards,nguyen2024trustworthy}. However, these models are restricted to text interpretability, whereas visual content also plays an effective role during a crisis. We don't have annotated explainable image patches, and annotating such rationale data is a challenging task.
\textit{In this paper, we try to bridge the existing research gaps and answer the following two questions --- (i). How do we extend the crisis classifiers' interpretability for images along with the text? and (ii). How can we use transfer learning to extract image rationales with the help of text rationales?}

\section{Data Collection}
\label{sec:data}

\begin{table}[tb]
\small
    \centering
     \setlength{\tabcolsep}{0.3em}
        \caption{List of humanitarian classes and data size}
    \label{tab:dataset}
    \vspace{-2mm}
    \begin{tabular}{|c|c|c|c|c|c|c|}
    \hline
      \multirow{2}{*}{Humanitarian class}  &  \multicolumn{2}{c|}{Train (70\%)} & \multicolumn{2}{c|}{Dev (15\%)} & \multicolumn{2}{c|}{Test (15\%)}\\
      \cline{2-7}
         & Text & Image & Text & Image &Text & Image\\
    \hline
    Infrastructure damage & 496 & 612 & 80 & 80 & 81 & 81\\
    \hline 
    Affected individuals & 70 & 71 & 9 & 9 & 9 & 9\\
    \hline 
    Rescue effort & 762 & 912 & 149 & 149 & 126 & 126\\
    \hline 
    Other relevant info & 1,192 & 1,279 & 239 & 239 & 235 & 235\\
    \hline 
    Not humanitarian & 2,743 & 3,252 & 521 & 521 & 504 & 504 \\
    \hline 
    Total & 5,263 & 6,126 & 998 & 998 & 955 & 955\\
    \hline 
    \end{tabular}
    \vspace{-1mm}
\end{table}

\subsection{Multimodal Dataset}

We use the publicly available CrisisMMD dataset~\cite{alam2018crisismmd} - the only multimodal benchmark with annotated humanitarian class labels. It contains tweet texts and images collected from seven natural disasters, primarily from English-speaking users, and supports multiple classification tasks. We focus on humanitarian categorization and follow the filtering and data-splitting setup of~\cite{ofli2020analysis}. Dataset statistics are reported in Table~\ref{tab:dataset}. Tweet texts and images share the same labels, providing complementary information essential for crisis situational awareness.


\subsection{Rationale Collection}

The original dataset contains only tweet-level class labels and no rationale annotations. We define text rationales as informative phrases in a tweet that explain its humanitarian label. A tweet may contain multiple non-consecutive phrases as text rationales. 
   We follow the setup proposed in~\cite{Nguyen2022towards} to annotate text rationales. Given the tweet and its humanitarian class, three annotators independently label rationale words. Word-level agreement is computed across annotators. We obtained a Fleiss-kappa~\cite{kraemer1980extension} agreement score of 0.71 among the annotators. The conflicts are solved by majority votes. 
   Examples of tweets and rationales are illustrated in Table~\ref{tab:tweet_example}. 
   
   We annotate rationales only for text. Image rationales are represented as patch-level heatmaps highlighting visual evidence for the prediction. Since annotating visual rationales (e.g., bounding boxes) is costly and ill-defined, we automatically derive image rationales by mapping from text rationales.

\begin{table}[!t]
     \centering
     \small
     \caption{Examples of tweets in different humanitarian classes. Rationales are highlighted.}
     \label{tab:tweet_example}
     \vspace{-2mm}
     \begin{tabularx}{\columnwidth}{|S|L|}
     \hline
        Class  &  Tweet \\
    \hline 
         Infrastructure damage & Hurricane Maria damage latest: \hl{100,000 homes without power} after storm batters Caribbean https://t.co/Xp1gjES7qv https://t.co/Z9SNQ0NaEd \\
    \hline 
        Affected individuals & Cyclone Mora : \hl{4 killed} in Manipur, 140 houses destroyed in Mizoram :: https://t.co/dMlEngezZ4 https://t.co/NIeyQeLkwt\\
    \hline 
        Rescue effort & \hl{\#RedCross assisting thousands impacted} by the devastating \#California \#Wildfires https://t.co/H0GdwJl6Ru https://t.co/Tj2XzbDTPm\\
    \hline 
    Other relevant info & RT @Wx\_Maps: \hl{All \#Tornado, Severe Thunderstorm, and Flash Flood warnings} for the month of August: https://t.co/FiRrxNPDZL\\
    \hline 
     \end{tabularx}
     
     \vspace{-3mm}
 \end{table}
  
\section{Methodology}
\label{sec:method}
\begin{figure*}
    \centering
    \includegraphics[scale=0.62]{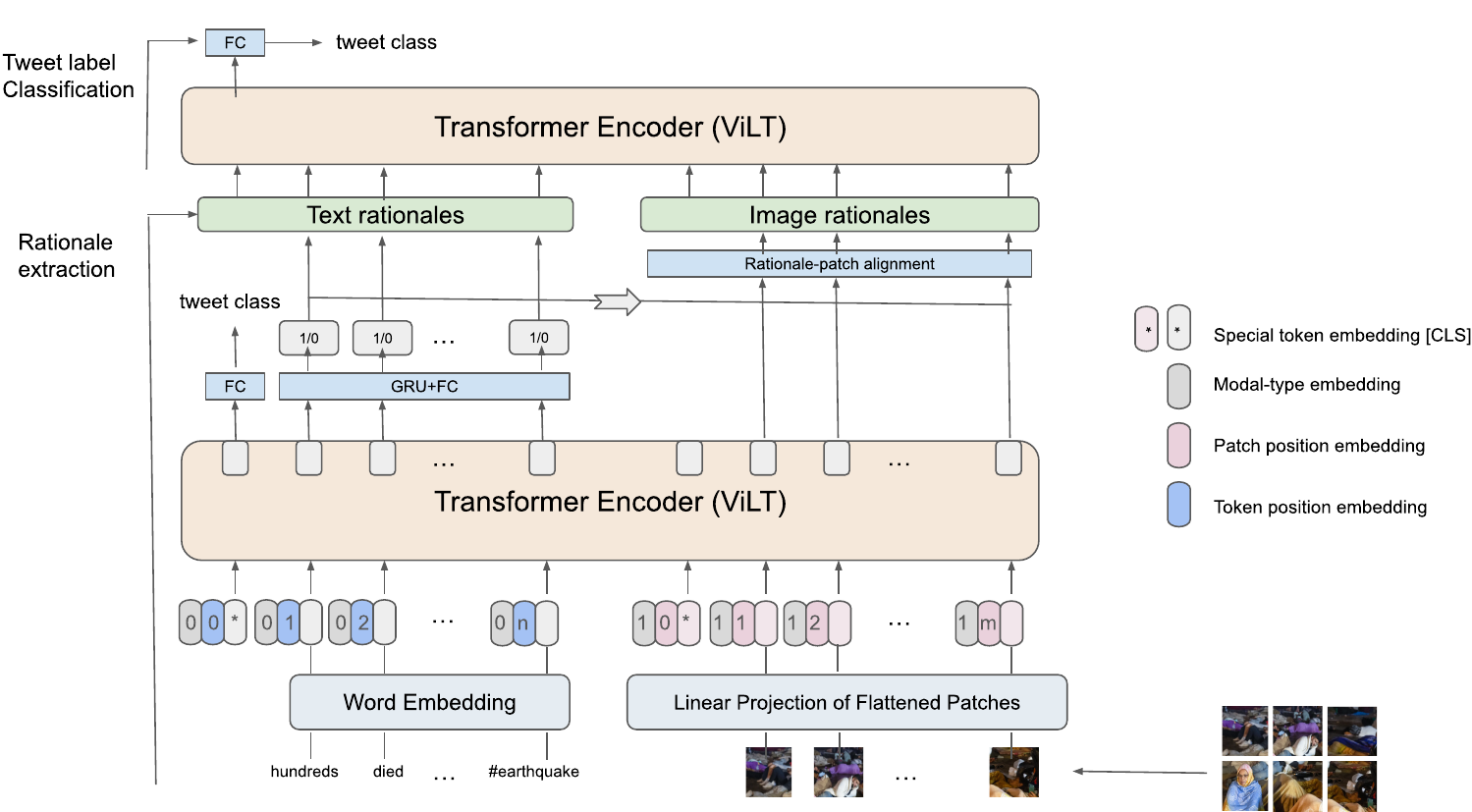}
    \caption{\method{} architecture. GRU and FC are Gated Recurrent Unit and Fully Connected Layer, respectively.}
    \label{fig:architecture}
    \vspace{-2mm}
\end{figure*}

\subsection{Overview}
 We aim to design a model that can accurately classify tweets into different humanitarian classes and extract faithful text and image rationales for model interpretability. Our model is a pipeline that consists of a rationale extraction and a tweet classification network (Figure~\ref{fig:architecture}). In the rationale extraction network, tweet texts and images are first preprocessed. Texts are tokenized into tokens; images are segmented and flattened into fixed-size patches. Then, we feed the data into a pre-trained transformer encoder to obtain corresponding tokens and patch embeddings. The primary goal at this step is to predict task-specific text rationales using supervised data. The rationales should be extracted according to tweet labels. Previous studies showed that learning rationales along with the task helps to learn better rationales~\cite{zhang2021explain}. Hence, we train our rationale extractor under a multi-task setup. The first task is to predict which text token is a part of the text rationales. The second auxiliary task is to predict humanitarian class labels of tweets. Then, we compute the alignment score between text rationale tokens and image patches to obtain an image rationale heatmap, which shows the important scores of image patches as supporting evidence for the tweet class labels.
 Next, we use the extracted text and image rationales as inputs to the tweet classification network for predicting tweet class labels. This step makes our \method{} interpretable by design, as the final class prediction is solely based on the rationales. 
 The detailed architecture of our approach is described as follows.
\subsection{Rationale Extraction}
\label{sec:rationale_extractor}
Our rationale extractor is a transformer-based network. It consists of an encoder and two rationale extractors that embed multimodal data input and extract text and image rationales for tweet labels. 
We formalize our tasks as follows. Given a set of tweets T, each $t\in T$ is represented as $t = <t_1, t_2, .., t_n, i_1, i_2, .., i_m>$, where $t_j$ and $i_k$ are tokenized text tokens and segmented image patches correspondingly. Our text rationale extractor predicts binary labels $r = <r_1, r_2, .., r_n>$, $r_j \in \{0, 1\}$ for each token $t_j$, where $r_j = 1$ if it is a part of text rationales. Then, the image rationale extractor assigns scores, so-called rationale heatmap, $h = <h_1, h_2, .., h_m>, h_k = [0,1]$ to specify how important each image patch $i_k$ is to the prediction.

\noindent\textbf{ViLT encoder}. In this study, we employ ViLT~\cite{kim2021vilt} to encode input data. 
Each tweet text is tokenized into a list of tokens $[CLS]$ $t_1$ $t_2$ .. $t_n$. An unknown word can be split into multiple tokens, and we save word-tokens correspondence. $[CLS]$ is a special token padded at the beginning of every tweet. The output ViLT embedding of $[CLS]$ token serves as a representation of the whole tweet text. Each image $i \in \mathbb{R}^{H\times W\times C}$ is segmented into a sequence of flattened 2D patches $[CLS]$ $i_1$ $i_2$ .. $i_m$, $i_i \in \mathbb{R}^{m\times (P^2\cdot C)}$, where $C$ is the number of channels, $(H, W)$ and $(P, P)$ are resolutions of the original image and image patches, respectively. $m = HW/P^2$ is the number of patches. Similar to text, $[CLS]$ token is prepended at the beginning, and its state at the output of ViLT encoder serves as image representation. The image and text inputs are concatenated into a sequence and fed to our encoder. The pre-trained ViLT encoder generates embeddings for each input token and image patch. Besides, ViLT also returns a pooler output embedding, which serves as the aggregated representation of the whole input multimodal sequence for downstream tasks. We chose ViLT as it is faster than previous Vision-Languague Pre-training (VLP) models. Besides, it returns pre-trained embeddings for every input token and image patch, enabling rationale extraction.

\noindent\textbf{Text rationale extractor}. The text rationale prediction is formulated as a binary token-level classification task. We convert the initial word-level rationale labels to token-level labels where split tokens have the same label as their original word. To encourage task-specific rationales, tweet label classification over humanitarian classes $l \in L$ (Table~\ref{tab:dataset}) is used as an auxiliary task, and both objectives are trained jointly.

We append a GRU (Gate Recurrent Unit) layer to capture the dependency between token embeddings, followed by a Fully Connected Sigmoid layer on top of ViLT encoder. The rationale extractor minimizes the following weighted loss:
\begin{gather}
    Loss_r = -\sum_{j=1}^n \frac{n}{n_j}BCE(y_j, p_j)
\end{gather}
where $y_j \in \{0, 1\}$ is the actual token rationale label, $p_j$ is the probability of a token $t_j$ to be rationale. $n$ and $n_j$ are the total number of tokens in the current tweet and the number of tokens having actual class labels $y_j$, respectively. BCE is the binary cross-entropy loss function. The weighted BCE helps to control the effect of the imbalanced number of rationale and non-rationale tokens in a tweet. This step predicts rationales at the token level. The word-level rationales can then be retrieved using max-pooling. 

For auxiliary tweet classification, we apply a fully connected Softmax layer to the ViLT pooler output and train it using a cross-entropy loss:
\begin{gather}
\label{eq:cross_entropy}
    Loss_l = -\sum_{l=1}^{L}y_llog(p_l)
\end{gather}
where $L$ is the set of humanitarian classes listed in Table~\ref{tab:dataset}. $y_l$ is the actual humanitarian class label of tweet $t$. $p_l$ is the probability of the tweet $t$ to be of label $l$.

Our text rationale extractor is then trained to minimize the following combined loss:
\begin{gather}
    Loss = Loss_l + \alpha Loss_r
\end{gather}
where $\alpha$ is used to control the effect of the two individual losses.

The ViLT encoder and text rationale extractor are trained end-to-end to learn text rationales and fine-tuned multimodal embeddings.

\noindent\textbf{Image rationale extractor}. Given a sequence of image patches $i = <i_1, i_2, .., i_m>$, this step identifies rationale heatmap $h = <h_1, h_2, ...h_m>$  to image patches. $h_k = [0, 1], h_k=0$ indicates the patch does not provide any supporting evidence for the tweet label. In contrast, the patch $h_k = 1$ provides an important explanation for the class label. We rely on predicted text rationales to generate the rationale heatmap $h$ of images. $h$ considers the similarity alignment between fine-tuned rationale token embeddings and image patch embeddings. Specifically, the alignment scores are first initialized with all zeros. Then, we compute the scores between embeddings of predicted text rationales and image patches using the inexact proximal point method (IPOT)~\cite{xie2020fast}. IPOT computes a soft matching between image patches and text tokens by approximating optimal transport, enabling accurate cross-modal alignment.
\begin{gather}
    h_k = max_{t_j, y'_j = 1}(IPOT(e_k, e_{t_j})), j = 1, 2, .., n
\end{gather}
where $e_k$ and $e_{t_j}$ are fine-tuned embeddings of image patches $k$ and text token $t_j$, correspondingly. $y'_j = 1$ indicates that token $t_j$ is predicted as rationale.

\subsection{Tweet classification}
\label{sec:classifier}
\begin{figure}
    \centering
    \includegraphics[scale = 0.37]{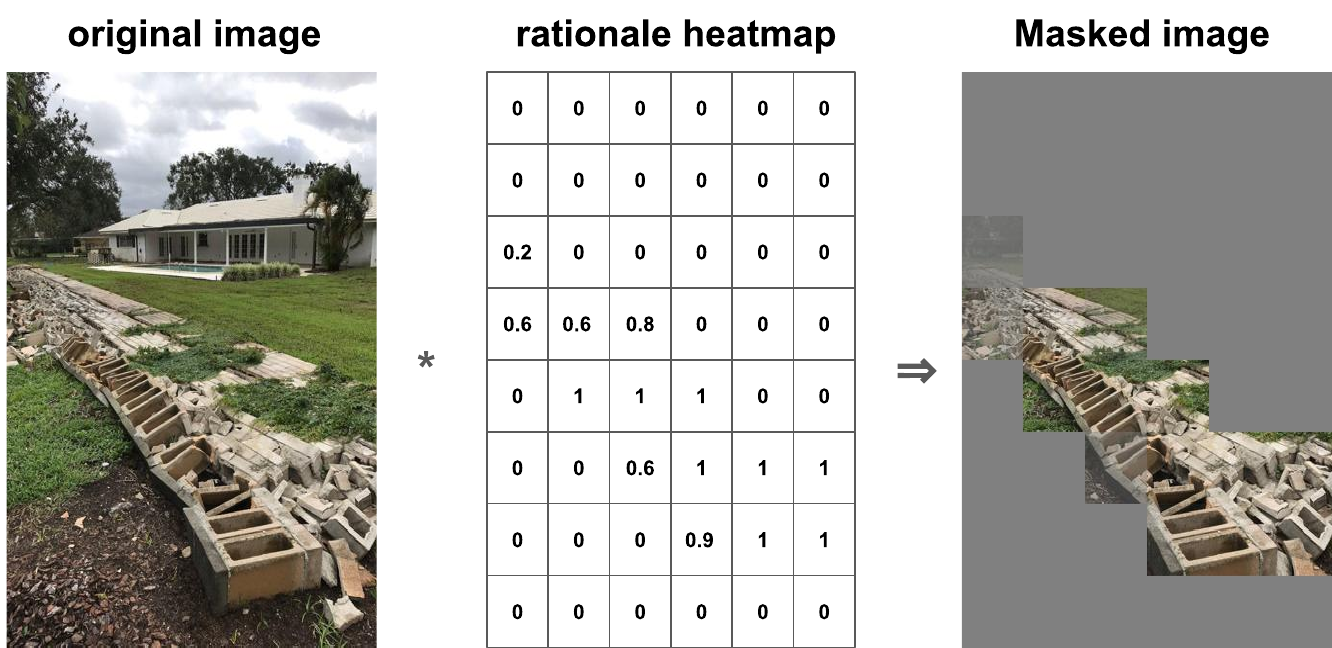}
    \caption{Example of an image masked by a rationale heatmap.}
    \label{fig:masked_image}
    \vspace{-6mm}
\end{figure}

In this phase, we use only predicted rationales as input data. First, we replace predicted non-rationale words with a special token $*$. Besides, we apply the rationale heatmap on image patches to blur non-rationale patches. For example, ``\textit{\#StormAlert9 possible tornado in Winter Park? Tree uprooted and whole wall collapsed . Weird}'' is the original text and ``\textit{* * * * * * * Tree uprooted and whole wall collapsed * *}'' is the text with masked non-rationale words. An example of a masked image by a rationale heatmap is illustrated in Figure~\ref{fig:masked_image}. This stage shows how well our extracted rationales support the model's classification outputs. We also employ ViLT as a multimodal encoder and apply a fully connected Softmax layer on top of the first token ViLT embedding to predict humanitarian classes of tweets. The classifier is trained to minimize the cross-entropy loss, similar to Equation~\ref{eq:cross_entropy}.

\section{Experimental Setup}
\label{sec:experimental_setup}

\subsection{Baselines}
We select baselines to cover representative modeling families in crisis informatics, including unimodal neural models, classical multimodal fusion approaches, and recent foundation models. This allows us to compare VLTCrisis against both task-specific and general-purpose systems.
Unimodal models establish the lower bound of what can be done with text or images alone. They are important to judge the requirements of including both modalities. 
 
\subsubsection{Unimodal Text Classification Models}

\begin{itemize}
    \item {\bf CNNT}~\cite{ofli2020analysis}: A CNN-based model for the classification of disaster-related tweet texts.
    \item {\bf RACLC}~\cite{nguyen2022rationale}: A rationale-aware interpretable learning approach for text classification of crisis events.
    \item {\bf JOMT}~\cite{zou-caragea-2023-jointmatch}: A semi-supervised text classification approach that relies on adaptive adjustment of class performance.
    \item \textbf{LLORA}~\cite{yin2024crisissense}: Instruction fine-tuned LLaMa~\cite{touvron2023llama} for tweet classification.
\end{itemize}

\subsubsection{Unimodal Image Classification Models}
\begin{itemize}
    \item \textbf{RNET}~\cite{petsiuk2018rise}: ResNet50 with RISE explanation.
    \item {\bf VGGI}~\cite{ofli2020analysis}: A VGG-based network for disaster-related image classification.
    \item {\bf DNET}~\cite{huang2017densely}: An unimodal image model pre-trained on ImageNet and fine-tuned on CrisisMMD image data.
    \item {\bf SCNN}~\cite{rahman2023learning}: Partial correlation-based deep visual representation learning approach for image classification.
\end{itemize}

\subsubsection{Multimodal Classification Models}
\begin{itemize}
    \item {\bf VCNN}~\cite{ofli2020analysis}: A multimodal deep learning method for classification of disaster events.
    \item {\bf FMLS}~\cite{sirbu2022multimodal}: A multimodal semi-supervised approach for tweet classification.
    
    \item {\bf GPT-4}: One-shot classification and text rationale extraction using GPT-4. We give a prompt and an instruction example to OpenAI API GPT-4~\cite{GPT} to obtain class labels and text rationales (See Appendix).
    \item \textbf{IncBERT}~\cite{basit2023natural}: A multi-modal classification model for disaster management. We apply LIME~\cite{ribeiro2016should} with 1000 perturbed samples to generate explanation.
        \item {\bf FCLIP}~\cite{mandal2024contrastive}: A fine-tuned crisis-related classification model.

\end{itemize}

\subsection{Evaluation Metrics}
\subsubsection{Task Performance}
Following previous works on human-
itarian classification (Nguyen and Rudra 2022a; Nguyen
et al. 2017), we evaluate the tweet classification task using
Macro-F1.
\subsubsection{Model Plausibility} Plausibility measures the extent to which humans agree to predicted rationales. We evaluate the
text rationale extraction performance using Token-F1 metric. It measures the similarity to groundtruth text rationales. Besides, we also perform human evaluations to judge whether humans think the extracted rationales are useful for them to identify class labels.

 We prepare two sets (Set1 and Set2) of text and image rationales that are sampled across five classes. Each set contains 10 text rationales and 10 image rationales. However, these sets are complementary i.e., corresponding image rationales of texts in Set1 are present in Set2 and vice versa. Ten annotators are recruited to evaluate each set ($2 \times 10=20$). Our annotators are graduate students or research scholars having English competency. They are introduced to the crisis-related information before doing this study. Altogether, we have 400 annotations ($2 \times 10 \times 20$).

The task is accomplished in two phases. In $Phase1$, we ask annotators two questions:- (i).~{\bf Q1.} Assign the text/image rationales to one of the class labels, (ii).~{\bf Q2.} Assess the difficulty of the task ($0$ indicates easy, $1$ difficult). In the $Phase2$, we provide both text and image rationales for each tweet and ask the same questions. This combined set contains 20 multimodal instances. The objective of this phase is to understand whether multimodal information helps users in reducing the difficulty and rectifying any errors that happened in $Phase1$.

\begin{table}[!tb]
    \centering
        \caption{Example of a tweet (X), rationales (R) or non-rationales only (X $\backslash$ R).}
    \label{tab:masking}
    \vspace{-2mm}
    \begin{tabularx}{\columnwidth}{l|X}
       X&  Pics from Iran 204 person killed and 1600 injured by \#earthquake \\
       
       \hline 
        R & * * * 204 person killed and 1600 injured * * \\
        \hline 
      X$\backslash$R & Pics from Iran * * * * * * by \#earthquake \\
        \hline

    \end{tabularx}

    \vspace{-3mm}
\end{table}

\subsubsection{Model Faithfulness}We evaluate how well predicted rationales support classification using comprehensiveness and sufficiency. Three input settings are considered: (i) $X$, the original multimodal input; (ii) $R$, rationales only, where non-rationale text is masked with `*', unimportant image patches are blurred using rationale heatmap), (iii).~$X\backslash R$: Predicted non-rationales only (rationale texts are replaced by `*', important image patches are masked out using non-rationale heatmap, where non-rationale heatmap = 1 - rationale heatmap).



Table~\ref{tab:masking} illustrates an example of such input text. Figure~\ref{fig:masked_image} shows an example of an image rationale. The two metrics are computed as follows: 

\textbf{Comprehensiveness}~\cite{DeYoung2020}: It measures the performance drop when we use only non-rationale input instead of the original input. The higher drop is better as it indicates that the most important parts of the inputs are captured by predicted rationales. 
\begin{gather}
   \label{eqn:comprehensiveness}
   \textit{ Comp. = Performance(X) - Performance(X $\backslash$ R)}
\end{gather}

\textbf{Sufficiency}~\cite{DeYoung2020}: It evaluates the change in model performance when we use rationales only compared to original inputs. The smaller sufficiency indicates that only rationales are sufficient for the model's prediction.
 \begin{gather}
    \label{eqn:sufficiency}
    \textit{Suff. = Performance(X) - Performance(R)}
\end{gather}

\subsection{Model Details and Hyperparameters}
\label{sec:config}

To train our \method{} model, we first pre-process tweets by converting texts to lowercase, removing user mentions and URLs. Images are normalized, the shorter edge is resized to 384, the longer edge is limited to under 640 while preserving the aspect ratio, and image patch size $P$ is set to 32 in the same way as the pre-processing step of ViLT. All experiments are run on an NVIDIA 1080Ti GPU using Pytorch and transformers packages.  We explored hyperparameters manually. The learning rate $[$3\text{e-}5, 1\text{e-}1$]$, batch size [8, 128], and number of maximum epochs [5, 20], $\alpha$ weight [1e-2, 1]. All the hyperparameters are tuned on the development set, where we select values resulting in the highest average Macro-F1 and Token-F1 on the development set. Finally, our \method{} is trained for a maximum of 10 epochs using early stop with patience = 3. We set the batch size to 8 and use AdamW optimizer~\cite{Loshchilov2019}, with a learning rate of 0.1. The GRU layer in the text rationale extractor has a hidden size of 128, $\alpha$ is set to 0.09. 

\section{Experimental Results}
\label{sec:results}
\subsection{Text-Image Alignment}
\label{subsec:alignment}

\begin{figure}
    \centering
    \includegraphics[scale=0.55]{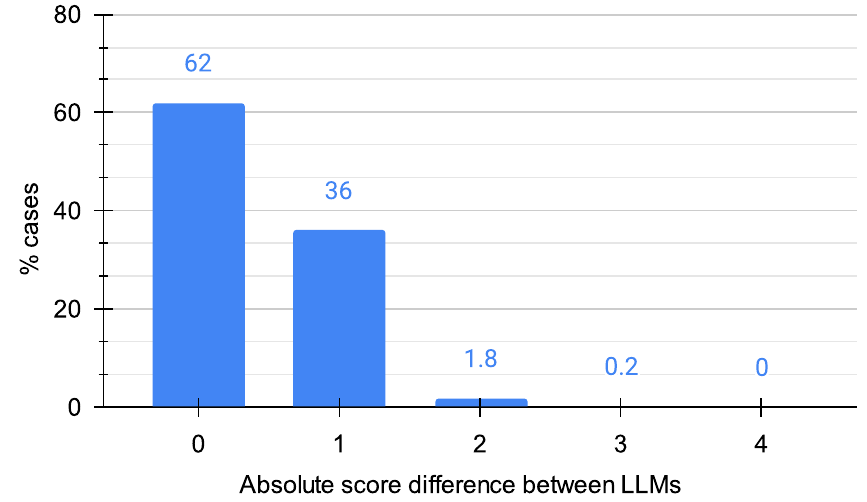}
    \caption{Histogram of absolute alignment score differences between Qwen-VL and LLaVA. A difference of 0 indicates exact agreement.}
    \label{fig:LLMs_scoreDiff}
\end{figure}

We validate our text-image alignment using two state-of-the-art open-source multimodal LLMs, QWen-VL~\cite{bai2025qwen2} and LLaVA~\cite{liu2023visual}. Qwen-VL achieves leading performance on recent multimodal benchmarks (e.g., MMBench, SEED-Bench), making it a strong proxy evaluator. LLaVA, while earlier, has become a widely recognized baseline in multimodal studies. Given a text-image pair of a tweet, we prompt the LLMs (see Appendix) to rate whether the text describes or aligns with the image on a Likert scale from 1 (no alignment) to 5 (perfect alignment). LLM-as-a-judge has shown its on-par performance with human evaluators and has been widely used for evaluation tasks~\cite{zheng2023judging}.

The evaluation shows that Qwen-VL and LLaVA give average alignment scores of 4.1 and 3.8, respectively. Qwen-VL was trained more extensively on multimodal grounding tasks, whereas LLaVA is primarily instruction-tuned and tends to be more conservative in its similarity judgments. However, we observe a high agreement between the two LLMs (Figure~\ref{fig:LLMs_scoreDiff}). 98\% of cases, scores returned by Qwen-VL and LLaVa have an absolute difference of 0 or 1, in which 62\% of cases,  the two models return an exact match.
\subsection{Task Performance}

\begin{table}[!t]
    \centering
    \footnotesize
     \setlength{\tabcolsep}{0.93em}
         \caption{Performance evaluation (Macro-F1) of \method{} (VLTC) and baseline models.}
    \label{tab:model_performance}
    \vspace{-2mm}
    \begin{tabularx}{\columnwidth}{|c|c|c|c|c|c|c|}
      \hline
  Text  & {CNNT} & {RACLC} & \multicolumn{2}{|c|}{LLORA}&\multicolumn{2}{|c|}{JOMT} \\ \cline{2-7}
   (T) & {0.470} & {0.640} & \multicolumn{2}{|c|}{0.660}&\multicolumn{2}{|c|}{0.630} \\
  \hline
  \hline
  Image  & {RNET} & {VGGI} & \multicolumn{2}{|c|}{DNET} & \multicolumn{2}{|c|}{SCNN} \\ \cline{2-7}
  (I)  & {0.595} & {0.623} & \multicolumn{2}{|c|}{0.605} & \multicolumn{2}{|c|}{0.630} \\
  \hline
  \hline
  T+I  & {VCNN} &  FMLS & GPT4 &{IncBERT} & FCLIP & \method{} \\ \cline{2-7}
    & 0.633 &  0.651 & 0.673 &0.804 &\textbf{0.834} &  0.822\\
    \hline
    \end{tabularx}
    \vspace{-3mm}
\end{table}

\textbf{Comparison with unimodal models.} We show the groundtruth-based evaluation of our experiments in Table~\ref{tab:model_performance}. 
\method{} outperforms four unimodal (text) models with large margins regarding the class label prediction task. Specifically, \method{} obtains an improvement of 16-35\% Macro-F1 score over CNNT, RACLC, LLORA, and JOMT. We observe that the fine-tuned large language model LLORA does not perform well on a relatively small, imbalanced dataset such as CrisisMMD.
Compared to the unimodal (image), \method{} achieves 19.2\%, 19.9\%, 21.7\%, 22.3\% higher Macro-F1 than SCNN, VGGI, DNET, and RNET, respectively.

\noindent\textbf{Comparison with multimodal models.} Multimodal models generally outperform unimodal methods, confirming the complementary nature of text and images. Apart from FCLIP, \method{} performs better than other baselines. 
GPT-4 correctly identifies class labels of tweets in 67.3\% cases. Tweets across different humanitarian classes are highly related, and many contain overlapping information. Hence, it leads to low performance of GPT. The result is in line with a previous work~\cite{huang2023is,imran2025evaluating}. IncBERT applies late-fusion embeddings for classification and performs worse than our \method{} by 1.8\%. FCLIP attains slightly better performance than \method{} but lacks interpretability, whereas our model is interpretable by design. We adopt ViLT rather than CLIP as the backbone because it provides token- and patch-level embeddings, which support the extraction of rationale for model interpretability.

Our \method{} misclassifies 18\% of tweets. We observe that failures are mainly due to the lack of data in the ``affected individuals'' class. Besides, \method{} also struggles to classify some tweets with mixed or overlapping information.  Some tweets from the `non-humanitarian' class also mention victims, infrastructure, and support that can confuse the model. We show some misclassification examples in the Appendix. 

Moreover, we evaluate model performance using a single modality and on individual events (see Appendix). Results show that our model performs generally well on all events. Both tweet texts and images are important for the
prediction. Using both text and image data results in a significantly
higher performance. Macro-F1 decreases notably from 82.2\% to 64.3\% and 69.1\%
when images or texts are not provided, respectively.

\subsection{Plausibility}
\label{sec:qualitative_performance}
\begin{figure*}
  
    \centering
    \small
    \begin{tabularx}{\textwidth}{XXX}
    \includegraphics[scale=0.38]{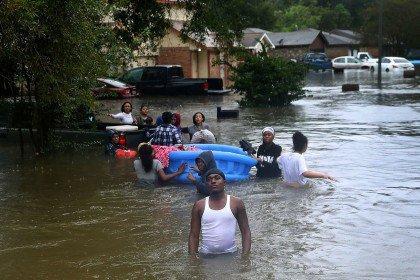}\vspace{-0.4cm}
&\includegraphics[scale=0.38]{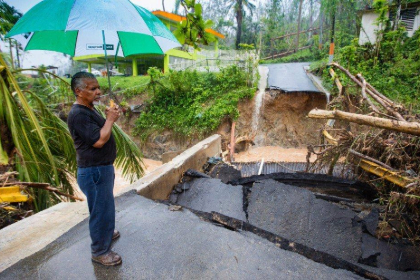}&\includegraphics[scale=0.38]{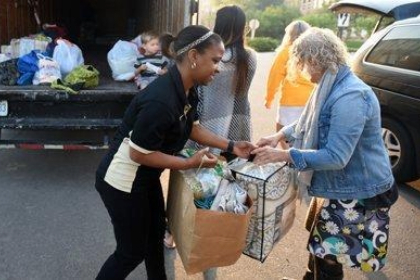}\\

         Storm Harvey flood victims face displaced alligators https://t.co/DvIeB2Jf3v https://t.co/JBTbHjmXIk\vspace{-0.4cm}&Guaynabo resident Efrain Diaz stands by a bridge washed out by rains carrying debris from Hurricane Maria. (CNN) https://t.co/M3aSh8qocx& Video of CU collecting donations for Hurricane Harvey. https://t.co/2Y5EXiA94w \#CUBoulder, \#CUBuffs, \#boulder https://t.co/8o68vyVdSFL\\
        \hline 
$\Downarrow$&$\Downarrow$&$\Downarrow$\\

\includegraphics[scale=0.38]{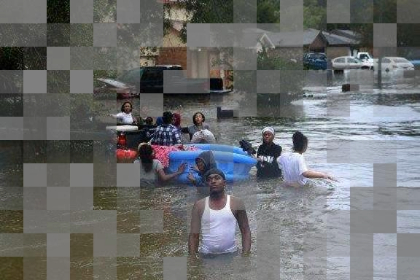} \vspace{-0.4cm} & \includegraphics[scale=0.38]{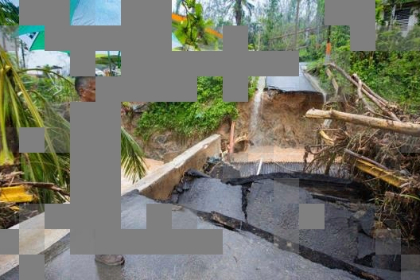}&\includegraphics[scale=0.38]{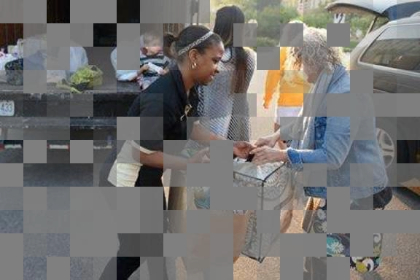} \\
        \hl{victims face displaced alligators} \vspace{-0.4cm}&\hl{a bridge washed out by rains carrying debris}& \hl{CU collecting donations} \\
        \hline 
      \textbf{ Class: Affected individuals}& \textbf{Class: Infrastructure damage }& \textbf{Class: Rescue effort}\\
         
    \end{tabularx}
       
     \caption{Examples of predicted labels and rationales. Rationale words are highlighted.}
    \label{fig:example1}
    \vspace{-2mm}
\end{figure*}

\begin{table}[!t]
    \centering
     \setlength{\tabcolsep}{0.4em}
       \caption{Token-F1 evaluation of text rationales}
    \label{tab:token-f1}
    \vspace{-2mm}
    \begin{tabular}{|c|c|c|c|c|}
    \hline 
   
   \textbf{Model} &GPT-4& IncBERT& RACLC  & \method\\
    \hline 
    \textbf{Token-F1} &0.552&0.611&0.811&0.826\\
    \hline 
    \end{tabular}
  
    \vspace{-4mm}
\end{table}

Comparing inherently interpretable models in Table~\ref{tab:token-f1}, \method{} obtains better Token-F1 than GPT-4 and RACLC, which means our method obtains higher agreement with human groundtruth of text rationales. Examples of tweets, along with the predicted images and text rationales, are shown in Figure~\ref{fig:example1}.

 Table~\ref{tab:human_evaluation} reports qualitative results of extracted rationales in two evaluation phases. In $Phase1$, the values for Q1 present the fraction of cases in which users can predict the correct class based only on the text or image rationales. Scores for Q2 show the fraction of cases users think the task is difficult. Similarly, $Phase2$ presents values for multimodal instances of two sets. For bot the sets, predicting classes from text rationales ($0.64/0.60$) or image rationales ($0.66/0.57$) is more difficult than their multimodal counterpart ($0.70/0.65$). Multimodal information helps to reduce the task difficulty by 11-13\%.

\begin{table}[tb]
    \centering
    \small
     \setlength{\tabcolsep}{0.6em}
       \caption{Human evaluation of \method{}. Values indicate the fraction of instances correctly annotated (Q1) and the difficulty to annotate (Q2).}
    \label{tab:human_evaluation}
    \vspace{-2mm}
    \begin{tabular}{|c|c||c|c||c|c|}
      \hline
      Phase & Set & \multicolumn{2}{c||}{Q1(Class Detection)} & \multicolumn{2}{c|}{Q2(Task Difficulty)} \\ \cline{3-6}

     &  & Text & Image & Text & Image \\
     \hline
     $Phase1$ & Set1 & 0.64 & 0.66 & 0.28 & 0.22 \\ \cline{2-6}
      & Set2 & 0.60 & 0.57 & 0.25 & 0.40 \\
      \hline
      \hline
    $Phase2$ & Set1 & \multicolumn{2}{|c||}{0.70} & \multicolumn{2}{|c|}{0.15} \\ \cline{2-6}
     & Set2 & \multicolumn{2}{|c||}{0.65} & \multicolumn{2}{|c|}{0.14} \\
      \hline
    \end{tabular}
\end{table}

 \textbf{\method{} vs RNET.} We compare image rationales extracted by \method{} and RNET qualitatively.  RNET is a post-hoc Randomized Input Sampling for
Explanation of Black-box Models, which showed better interpretability ability than some other explanation approaches~\cite{petsiuk2018rise}. RNET produces saliency heatmaps or importance heatmaps of image pixels for classification explanation.
For fair comparison with our method, we keep the top 25\% most important pixels of RNET as rationales, other pixel values are set to zeros. Similarly, 25\% of the most important patches with the highest rationale heatmap values by \method{} are kept as rationales, and the remaining patches are blacked out. Then, we sample 20 images that are correctly classified by both the method and extract image rationales for human evaluation. It turns out that our image rationales are more helpful in identifying class labels. More specifically, when we give image rationales and ask users for class labels, users can correctly identify humanitarian labels of tweets with an accuracy of 74\% and 62\% for \method{} and RNET, respectively. Moreover, given two sets of image rationales and corresponding class labels, users also chose \method{} over RNET. More details of the evaluation and results are shown in Appendix.

\subsection{Faithfulness}

\begin{table}[tb]
    \centering
      \caption{Model Comprehensiveness and sufficiency.}
    \label{tab:faithfulness_rise}
    \vspace{-2mm}
    \begin{tabular}{|c|c|c|}
    \hline
   Method &Comprehensiveness$\uparrow$&Sufficiency$\downarrow$\\
    \hline
\hline
    \method & 0.514 & 0.072\\
    \hline 
    RNET & 0.174 & 0.352 \\
    \hline
    RACLC &0.475&$-0.001$\\
    \hline
    \end{tabular}
   
\end{table}
We evaluate faithfulness using comprehensiveness and sufficiency for \method{} and interpretable baselines (RACLC, RNET). As shown in Table~\ref{tab:faithfulness_rise}, \method{} achieves high comprehensiveness and low sufficiency, indicating that masking rationales causes a large performance drop, while rationales alone are sufficient for prediction.

Compared to RACLC, our method obtains higher comprehensiveness and higher sufficiency. We observe that RACLC fails to cover some rationale tokens in the input texts, hence, it has lower comprehensiveness compared to \method{}.
Besides, top pixels returned by \method{} are more comprehensive and sufficient for prediction than RNET. By removing less important pixels, \method{} performance decreases by 7.2\%. Meanwhile, RNET performance drops significantly by 35.2\% when less important pixels are zeroed out. In contrast, \method{} performance decreases significantly (51.4\%) compared to RNET (17.4\%) when important pixels are masked. Overall, the results show \method{} is more faithful than RNET.

\subsection{Adaptability to Future Events}
We evaluate the model generalizability on the newly collected multimodal DMD dataset~\cite{mouzannar2018damage}, which was introduced for damage identification. We randomly sample  60 tweets, apply our model to obtain class labels, text rationales, image rationales, and distribute the tweets into two evaluation sets. Ten human annotators are recruited to evaluate each set independently by answering four questions.
\begin{itemize}
\item \textbf{Q1}. Given the original tweet and predicted label. Answer [Yes/No] whether the label is correctly assigned.
\item \textbf{Q2}. Answer [Yes/No] whether the extracted text rationale accurately supports the label.
\item \textbf{Q3}. If Q2 is responded with "Yes". Answer [Yes/No] whether the text rationale contains only relevant rather than redundant words.
\item \textbf{Q4}. Answer [Yes/No] whether image rationales align well with the text rationales. 
\end{itemize}

As shown in Table \ref{tab:new_event_adaptability}, the model performs strongly in Set 1 (>80\% across all criteria). Set 2 exhibits lower agreement for label correctness and rationale accuracy (Q1: 0.72, Q2: 0.68), but comparable rationale relevance and multimodal alignment (Q3: 0.85, Q4: 0.83). Annotators reported that the mixed information of tweets in set 2 caused confusion. Notably, the high Q3 scores indicate that, among instances with correct text rationales, approximately 85\% of explanations were judged non-redundant in both sets. Besides, across both evaluation sets, image rationales showed consistently strong alignment with their corresponding text rationales (0.83 in both sets). Overall, our model produces generally accurate labels and coherent multimodal rationales, but there is still scope to improve the redundancy of text rationales.
\begin{table}[tb]
    \centering
       \caption{Fraction of "Yes" responses by human judgements.}
    \label{tab:new_event_adaptability}
    \vspace{-2mm}
    \begin{tabular}{|c|c|c|c|c|}
    \hline 
        Set &  Q1 & Q2 & Q3 & Q4\\
    \hline
       Set1  & 0.85 & 0.82 & 0.85 & 0.83\\
    \hline
        Set2 & 0.72&0.68&0.85&0.83\\
    \hline
    \end{tabular}
    \vspace{-5mm}
\end{table}
\section{Concluding Discussion}
\label{sec:conclusion}
In this paper, we propose a multimodal interpretable approach to classify crisis-related tweets. Existing interpretable approaches mostly focus on text. However, images play a crucial role in understanding the severity of the situation. Prior works on crisis-related image classification tasks did not pay attention to image explanations. In this work, we take a step towards that, extracting text and image rationales, and developing a classification framework using those rationales. We explore the power of transfer learning to get image rationale patches based on the annotated text rationales. This could help to reduce annotation efforts and learn good-quality image explanations from their text counterpart.

\begin{acks}
This work was supported in part by \grantsponsor{}{the Scheme for Promotion of Academic
and Research Collaboration (SPARC)}{} (No. \grantnum{}{SPARC/2024-2025/ADVCOM/P3764}), and \grantsponsor{}{Microsoft Academic Partnership Grant 2023}{} (No. \grantnum{}{7581365}). Koustav Rudra is a recipient of \grantsponsor{}{the DST-INSPIRE Faculty Fellowship}{} (\grantnum{}{[DST/INSPIRE/04/2021/003055]} in the year 2021 under Engineering Sciences).

\end{acks}



\bibliographystyle{ACM-Reference-Format}
\bibliography{reference}
\subsection{Appendix}

\noindent \textbf{Misclassification}. We show some misclassification example in Table~\ref{tab:misclassification}.
\begin{table*}[h]
    \centering
    \small
     \caption{Misclassification Examples}
    \label{tab:misclassification}
    \vspace{-2mm}
    \begin{tabularx}{\textwidth}{|W|M|M|}
    \hline 
       \textbf{Tweet}  &  \textbf{Predicted Class} & \textbf{Groundtruth}\\
       \hline 
       Harvey victims' nerves fray as days drag in Houston shelter [URL] [URL]	  & Affected individuals & Rescue effort\\
        \hline 
    \#CycloneMora As usual \#Buddhist Monks are door to door to their peoples [URL]&Affected individuals &not humanitarian\\
    \hline
    Fundraiser for hurricane Irma and Harvey with Women Beyond Survival&other relevant info& rescue effort\\
    \hline
    If you’re looking for a way to help, my \#Church has a few good options. \#hurricane \#harvey \#relief \#cottonwoodfamily [URL]& other relevant info& rescue effort \\
    \hline 
    9/18/ Post Irma information. Please continue to handle storm waste safely, In the event of an emergency dial 911… https://t.co/3w8GJkT0EA&rescue effort&other relevant info\\
    \hline
    Labrador Helping Mexicoâ€™s Earthquake Efforts Earns Presidential Thanks [URL] [URL]& rescue effort & not humanitarian\\
    \hline 
    Flooded \#Cars Could Flood \#UsedCar Market [URL] @WardsAuto \#Irma \#Harvey [URL]&infrastructure damage&not humanitarian\\
    \hline
    \end{tabularx}
\end{table*}

\begin{table}[!ht]
    \centering
    \setlength{\tabcolsep}{0.3em}
    \footnotesize
     \caption{\method{} performance on individual events for infrastructure (INF), rescue (RES), affected individuals (AFF) and other relevant 
 (OTH). Best values are in bold. Red boxes indicate where method falls short. NA if the class is absent.}
    \label{tab:event_wise_performance}
    \vspace{-2mm}
    \begin{tabular}{|c|c|c|c|c|c|c|c|c|}
    \hline 
        \multirow{2}{*}{Event}  & \multicolumn{4}{c|}{Macro-F1}&\multicolumn{4}{c|}{Token-F1}\\
        \cline{2-9}
          & AFF& INF & RES & OTH & AFF&INF & RES & OTH \\
        \hline 
    Hurricane Irma &{0.00}&0.89 &0.75 &0.87&  1.00&0.768&0.795 &0.761 \\
    \hline 
      Hurricane Maria&{0.00} &0.93 &0.71 &0.90 &0.33 & \textbf{0.811} &0.787 & 0.639 \\
      \hline 
       Hurricane Harvey &{0.00}&0.93 &\textbf{0.90 }& 0.89 & 0.96 &0.685 & 0.852&  0.758 \\
    \hline 
          Srilanka Floods &NA&1.00 &0.67 &\textbf{1.00 }& NA&0.667& 0.857&\textbf{ 0.941} \\
    \hline 
    Mexico Earthquake &NA &0.91 &0.81 & {0.53}&NA  &0.747 & 0.769& 0.730 \\
    \hline 
    Iran-Iraq Earthquake  &0.86&\textbf{1.00} & {0.50} & {0.40}  & 0.64& 0.833 &\textbf{1.000} & 0.667 \\
    \hline 
    California Wildfire & NA&0.87& 0.67&0.62 &NA & 0.751& 0.801& 0.801\\
    \hline 
    \end{tabular}
    
\end{table}
\noindent\textbf{\method{} performance across events}.
Performance of our \method{} for different events is shown in Table~\ref{tab:event_wise_performance}. Our method performs generally well across events. In a few cases, performance falls short, such as `other relevant information' class on Mexico and Iran-Iraq earthquakes. The low Macro-F1 scores are mainly due to the small number of training instances for those events. \method{} achieves 86\% Macro-F1 for `affected individuals' class on the Iran-Iraq earthquake. For other events, there are either no or a few instances of the class, and our method fails to detect them. Gathering more training instances could be one way to address the problem. We would address these limitations using self-supervised contrastive learning in the future.

\begin{table}[!t]
    \centering
     \caption{\method{} Macro-F1 under different setups.}
    \label{tab:role_text_image}
    \vspace{-2mm}
    \begin{tabular}{|c|c|c|}
    \hline 
        \method{}  & \method{} (Text) &  \method{} (Image)  \\
    \hline 
     0.822 & 0.643 & 0.691\\
   \hline
    \end{tabular}
   
    \vspace{-5mm}
\end{table}

\noindent{\textbf{Uni-modal performance}}
In this part, we want to observe how different modalities
(i.e., text and image) influence VLTCRISIS performance.
We feed the VLTCRISIS tweet classification phase with
three input settings: Original texts and images, texts only, and images only. When feeding texts only, we black out the input images. Similarly, all words in the input texts are replaced by a special symbol, `*', for the model setup with image only. It can be seen from Table~\ref{tab:role_text_image} that both tweet texts and images are important for the prediction. Using both text and image data results in a significantly higher performance. Texts and images together provide more useful information to identify class labels, especially for classes of small sizes. Macro-F1 decreases notably from  82.2\% to 64.3\% and 69.1\% when images or texts are not provided, respectively. Besides, we also observe that the model using only image data achieves better Macro-F1 than text data. Specifically, \method{} (image) obtains $\sim$4\% better performance than \method{} (text).

\noindent{\textbf{Model Plausibility: VLTCRISIS vs RNET.}} We sample twenty images that are correctly classified by both our method (\method) and RNET. Images are sampled from all five humanitarian classes. Users are then asked to evaluate our image rationales. Figure~\ref{fig:example1} (Section \ref{sec:qualitative_performance}) shows the examples of rationale text and images.

We recruited the same twenty annotators and randomly distributed them into two groups ($G1$ and $G2$), each containing 10 members. The first group ($G1$) performs the annotation of our method while $G2$ evaluates the images for RNET. The task was explained to them before the annotation process. We have altogether $400$ annotations, i.e., $10 \times 20 = 200$ for each $G1$ and $G2$. 
The task is accomplished in two phases.
\begin{itemize}
    \item In $Phase1$, we provide image rationales to the annotators and ask them the following question: ({\bf Q}).~Assign the image to a humanitarian class based on the provided rationales. Annotators of $G1$ and $G2$ perform the task independently. After completion of $Phase1$, we start $Phase2$. 
    \item In $Phase2$, we give all the annotators ($20$) the following information - (i).~image rationales of \method, (ii).~image rationales of RNET, and (iii).~Ground truth class label. They are asked to answer the following question ({\bf Q}).~Given two different image rationales and corresponding humanitarian class label, identify which one is more helpful in deciding the class label. 
\end{itemize}

{\it Note, we do not reveal the method names to the annotators in both phases to remove annotation bias from the evaluation.} We also distribute annotators in two groups in $Phase1$ to ensure the fair evaluation of $Phase2$ because if the same annotator is exposed to both the image rationales in $Phase1$, she gets biased in $Phase2$.
\begin{table}[tb]
    \centering
   \caption{Human evaluation of \method{} and RNET. Values indicate the fraction of instances correctly annotated ($Phase1$) and preference of one method ($Phase2$).}    \label{tab:human_evaluation_RNET}
   \vspace{-2mm}
      \begin{tabularx}{\columnwidth}{|S|L|}
      \hline
      \multicolumn{2}{|c|}{Phase1} \\ \hline
      {\bf Method} & {\bf Fraction of cases where class label correctly identified} \\ \hline
      \method & 0.74 \\ \hline
      RNET & 0.62 \\ \hline \hline
      \multicolumn{2}{|c|}{Phase2} \\ \hline
      {\bf Fraction of cases \method{} is useful} & {\bf Fraction of cases RNET is useful} \\ \hline
      0.81 & 0.19 \\
      \hline
    \end{tabularx}
\end{table}

Table~\ref{tab:human_evaluation_RNET} shows the scores for both phases. Our proposed method (\method) turns out to be more helpful for annotators in identifying the class label. 
\method{} helps annotators to decide class label 20\% more cases over RNET correctly. This is also verified in $Phase2$ where annotators explicitly chose \method{} over RNET. On average, 80\% cases \method{} is more helpful than RNET in identifying class information. This justifies the claim that  `interpretable by design' approach turns out to be more helpful than posthoc approaches. Further, we perform a quantitative evaluation to verify the findings over a larger test set.

\noindent\textbf{GPT-4 Multimodal One-Shot Classification}. We give the following prompt and an instruction example to GPT-4 to obtain class labels and text rationales.\\

\fbox{
 \begin{minipage}{0.93\linewidth}
 \noindent Given the following tweet text and image, assign the tweet to one of the labels: [LABEL LIST]. Extract short phrases from the original tweet as explanations for the assigned label. \\
 Tweet: \{tweet\}\\
 Image: \{image\}\\
 Example:\\
      \noindent \textbf{Tweet}: \textit{with a \$ 20 million fund in his sights , j . j . watt delivers relief to texas harvey victims}\\
      \textbf{Image}: [IMAGE\_URL]\\
        \textbf{Label}: rescue\_volunteering\_or\_donation\_effort
    
       \noindent \textbf{Rationales}:
        \begin{itemize}
            \item[+]20 million fund 
            \item[+]watt delivers relief
        \end{itemize}
    \end{minipage}
}
\\

\noindent\textbf{Text-Image Alignment}. We give the following prompt to Qwen-VL and LLaVA to obtain alignment scores between text-image pairs:\\

\noindent\fbox{%
  \begin{minipage}{0.93\linewidth}
You are an evaluator. \\
Text: \{text\}\\
Image: \{image\}\\
Question: Does the text describe or align with the image?
Answer with:
\begin{enumerate}
    \item Score: [Give a similarity score from 1 (no alignment) to 5 (perfect alignment)]
\item Explanation: [a short justification] 
\end{enumerate}
\end{minipage}
}\\




\end{document}